\documentclass[10pt]{article}
\usepackage[margin=0.78in]{geometry}
\usepackage[T1]{fontenc}
\usepackage{lmodern}
\usepackage{microtype}
\usepackage{amsmath,amssymb}
\usepackage{booktabs}
\usepackage{tabularx}
\usepackage{array}
\usepackage{enumitem}
\usepackage{xurl}
\usepackage[hidelinks]{hyperref}
\hypersetup{
  pdftitle={AIREP: A Protocol for Per-Decision Evidence in AI Runtime Governance},
  pdfauthor={Ali Toygar Abak},
  pdfsubject={Per-decision cryptographic evidence for AI runtime governance},
  pdfkeywords={AI governance, runtime evidence, agent governance, auditability, cryptographic provenance, canonical JSON, interoperability, AI accountability}
}
\usepackage{graphicx}
\usepackage{parskip}
\setlist{nosep,leftmargin=*}

\newcommand{\code}[1]{\texttt{#1}}
\newcommand{\sha}[1]{\texttt{#1}}

\title{\textbf{AIREP: A Protocol for Per-Decision Evidence in AI Runtime Governance}}
\author{Ali Toygar Abak\\Independent Researcher\\\href{https://orcid.org/0009-0002-3718-4010}{ORCID: 0009-0002-3718-4010}}
\date{16 September 2026}

\begin{document}
\maketitle

\begin{quote}\small
\textbf{Manuscript scope.} This paper describes the protocol architecture and reports a dated implementation/evidence snapshot; it is not a release changelog. The exact byte-level construction evaluated here is AIREP wire version 0.2. The released implementation target is \code{v0.2.0-beta.1}; post-release integration evidence is pinned separately through 16 September 2026. AIREP remains experimental and is not a ratified standard, certification scheme, or regulatory-compliance claim.
\end{quote}

\begin{abstract}
Runtime-governance evidence is often collapsed into one audit event even though materially different events may have occurred: a governance decision may have been made, an instruction may have been dispatched or received, an action may or may not have executed, and a resulting state may or may not have been observed. Treating these events as interchangeable makes it difficult to determine what an audit record actually establishes.

This paper presents AIREP, a vendor- and model-independent protocol for per-decision AI runtime evidence. The current evaluated wire model separates evidence into four artifact families: Decision, Control, Execution, and Effect. Each artifact uses a closed core schema, explicit identities and digests, declared scope limitations, RFC 8785 canonical JSON, domain-separated SHA-256 hashing, and pure Ed25519 signatures. A three-level assurance model distinguishes structural and hash consistency (AIREP-Core), verifier-accepted authorship (AIREP-Authenticated), and independently anchored chain-head freshness and non-truncation relative to an accepted witness (AIREP-Witnessed). These classes concern provenance, integrity, and freshness; they do not establish the truth of the reported event.

A structured reconciler evaluates relationships among supplied artifacts while preserving failure, missing evidence, unevaluated prerequisites, and indeterminate outcomes as distinct states. The released beta implementation includes a four-family first-party producer, Python and Node reference-verification paths, adversarial and lifecycle corpora, and a reproducible release-validation runner. Post-release first-party integration exercises test a source-pinned mapping of the public Hermes approval/runtime source model using deterministic synthetic fixtures and preserve LightEval output through an experimental companion evaluation profile; neither exercise is an external-adoption or interoperability claim. Independent implementation evidence exists for a v0.1.2 producer and, separately, a v0.2 consumer/verifier, but the measurements target different frozen version bases and therefore do not establish same-version producer-to-consumer interoperability. AIREP remains experimental at the evidence snapshot reported here.
\end{abstract}


\section{Introduction}

AI systems increasingly make or mediate decisions that have consequences beyond text generation. A governed runtime may release an output, block it, defer it to a human, issue an external instruction, suppress a tool action, or attempt to stop an already active process. Once a decision crosses a process, trust, or administrative boundary, several questions become distinct:

\begin{enumerate}
  \item What governance decision was made?
  \item What instruction was actually dispatched?
  \item Did a receiving boundary report receipt?
  \item What action did an executor report attempting?
  \item Did the recorded executed action match the recorded authorized action?
  \item What state, if any, was subsequently observed?
  \item What assurance can an independent verifier place in each report?
\end{enumerate}

Adjacent technologies answer different parts of this problem. Model Cards document model characteristics and limitations rather than individual runtime events~\cite{modelcards}. OpenTelemetry standardizes common telemetry semantics~\cite{otel}. OpenID AuthZEN standardizes communication of authorization requests and decisions between Policy Enforcement Points (PEPs) and Policy Decision Points (PDPs)~\cite{authzen}. in-toto and SCITT address cryptographic provenance and transparency for signed statements and supply-chain artifacts~\cite{intoto,scitt}. Governance frameworks such as NIST AI RMF 1.0 and the EU AI Act establish broader risk-management and record-keeping contexts~\cite{nist,euai}. These roles are complementary, but they are not interchangeable.

AIREP addresses a narrower question: how can evidence about an individual governed AI-runtime decision and its subsequent control, execution, and observed-effect lifecycle be represented so that a third party can recompute its cryptographic bindings and see exactly where the evidence stops?

The first AIREP paper described the v0.1 line, centered on one decision-shaped core record~\cite{airepv1}. Subsequent protocol work introduced breaking changes rather than additive extensions. The current wire model separates lifecycle stages, closes the core schema, fixes byte-level integrity constructions, narrows assurance semantics, and adds structured reconciliation over a finite evidence set. Table~\ref{tab:changes} summarizes selected changes.

\begin{table}[ht]
\centering
\small
\caption{Selected changes from the v0.1 line to the current evaluated wire model.}
\label{tab:changes}
\begin{tabularx}{\textwidth}{@{}p{0.19\textwidth}X X@{}}
\toprule
\textbf{Property} & \textbf{v0.1 line} & \textbf{Current wire model (0.2)} \\
\midrule
Lifecycle model & One core decision record; later lifecycle facts represented through profiles & Four sibling artifact families: Decision, Control, Execution, Effect \\
Identity & Relative predecessor binding; no required chain or record identity & Required \code{chain\_id}, \code{record\_id}, and monotonic \code{sequence} \\
Integrity construction & In-place JCS rule, with historical reference-package alignment gaps & Frozen in-place JCS construction plus version/type domain separation and exact signature preimages \\
Core extensibility & Several open subobjects & Closed core; namespaced \code{profiles} is the extension surface \\
Portable signature baseline & Open algorithm label; v0.1 admitted HMAC at the former Verified level & Pure Ed25519 mandatory for the current suite; verifier binding selects cryptographic behavior \\
Assurance ladder & Core / Verified / Trusted & AIREP-Core / AIREP-Authenticated / AIREP-Witnessed with an explicit non-truth boundary \\
Lifecycle correlation & Primarily record/profile-local & Explicit cross-artifact references and structured reconciliation \\
\bottomrule
\end{tabularx}
\end{table}

The contributions of this revision are:

\begin{enumerate}
  \item \textbf{Lifecycle separation.} Decision, Control, Execution, and Effect are sibling evidence families, so evidence for one stage cannot silently be promoted into another.
  \item \textbf{Deterministic wire processing.} Raw JSON admission, RFC 8785 canonicalization, domain-separated hashing, and exact Ed25519 preimages are specified as byte-level behavior.
  \item \textbf{Bounded assurance semantics.} Provenance, integrity, freshness, failure, withholding, and caveats are distinct from truth and from lifecycle completeness.
  \item \textbf{Evidence-preserving reconciliation.} Missing, invalid, conflicting, or indeterminate evidence remains visible rather than becoming an implicit pass.
  \item \textbf{Extension without semantic collapse.} Namespaced profiles can carry companion evidence, including evaluation evidence, without creating a new lifecycle family or an assurance uplift.
  \item \textbf{Measured implementation status.} The paper reports released first-party validation, first-party integration exercises against external systems, and external implementation evidence at their actual version and independence boundaries.
\end{enumerate}

\subsection{Manuscript scope and protocol status}

This manuscript intentionally separates the research contribution from software release numbering. The conceptual claims concern evidentiary-stage separation, deterministic binding, bounded assurance, and reconciliation semantics. Exact field sets and byte preimages are necessarily versioned protocol details. The technical construction evaluated in this paper is wire version \code{0.2}.

The released implementation target used for the reference validation is \code{v0.2.0-beta.1}, published 9 September 2026~\cite{beta}. Its annotated tag resolves to commit \sha{8a6c01ecce457aa94330c0ed7219e4c56ebfe771}. The manuscript also cites a later repository snapshot, commit \sha{32f2ded6431025bf2737a7abe7152f24ccf72679} (16 September 2026), to report post-release integration work without retroactively changing what the beta tag contained~\cite{snapshot}. AIREP v0.1 remains a separate frozen line under its own rules. A v0.1 artifact cannot be relabelled as v0.2.

The beta is experimental. It is not stable interoperability certification, a security certification, regulatory compliance, or standards-body endorsement.

\section{Evidence Model and Threat Boundary}

AIREP's central design choice is to separate what event a record reports from what assurance the record itself earns. The protocol therefore adopts the following non-implications:

\[
\text{Decision established} \not\Rightarrow \text{Control delivered},
\]
\[
\text{Control delivered} \not\Rightarrow \text{Execution occurred},
\]
\[
\text{Execution occurred} \not\Rightarrow \text{Intended effect observed},
\]
\[
\text{Effect observed} \not\Rightarrow \text{Underlying decision correct}.
\]

These are not merely operational caveats. They are the reason the current wire model has four artifact families.

\subsection{Evidence is not truth}

Cryptographic integrity can establish properties of a record without establishing the truth of its content. A producer controlling a valid signing key can emit a well-formed, correctly signed false report. It can also omit relevant events outside the supplied evidence set. AIREP therefore does not treat a valid signature, chain, or witness as truth assurance.

A useful decomposition is:

\begin{table}[ht]
\centering
\small
\begin{tabularx}{\textwidth}{@{}p{0.24\textwidth}X@{}}
\toprule
\textbf{Layer} & \textbf{Bounded claim} \\
\midrule
Artifact validity & Accepted structure and internal cryptographic consistency \\
Authorship & Signature validity under a verifier-accepted producer binding \\
Freshness/non-truncation & Head anchoring relative to an accepted independent witness and freshness policy \\
Reconciliation & Relationships evaluated over the finite admitted evidence set supplied to the reconciler \\
Real-world truth & Not established by AIREP \\
\bottomrule
\end{tabularx}
\end{table}

\subsection{Explicit scope}

Every artifact carries a \code{scope} object with both \code{covers} and \code{does\_not\_cover}. This makes the producer's asserted evidentiary boundary integrity-protected. The field is not a mechanism for turning an incorrect claim into a correct one: scope statements remain producer assertions and inherit the malicious-producer limitation.

\section{Protocol Artifact Model}

Each family reports one lifecycle stage. Higher assurance does not change the stage and does not prove that the reported event occurred in the physical or external world.

\subsection{Decision}

A Decision artifact represents a governance decision over an input and stated policy basis. Its family fields include governed-input reference and digest, a claim with a nonempty basis array, a directive, an output reference and digest, and evidence references. The directive verb is one of \code{release}, \code{block}, \code{defer}, \code{redact}, \code{escalate\_to\_human}, or \code{kill}.

A Decision is neither a delivery receipt nor proof that downstream enforcement occurred. It is also important to distinguish \emph{policy basis} from an exact historical policy snapshot. The current Core can name policy basis and bind policy evidence by digest where supplied, but it does not require a version-pinned policy snapshot sufficient to replay the exact historical policy semantics. Cryptographic recomputation of an AIREP artifact therefore must not be described as guaranteed semantic re-evaluation under the policy that happened to exist at the original decision time.

\subsection{Control}

A Control artifact reports a boundary-side observation concerning an instruction. It binds the Decision, an instruction identifier and digest, an \code{authorized\_action\_digest}, the control event, boundary side, and authority declaration. The event is \code{dispatched}, \code{received}, or \code{delivery\_failed}; the boundary side is \code{issuer} or \code{receiver}.

Dispatch and receipt are deliberately asymmetric. An issuer can report that it dispatched an instruction, but that report is not receiver-side evidence of receipt. Likewise, the required \code{authority.writable\_by\_controlled\_system} field is a producer declaration, not independently verified authority.

\subsection{Execution}

An Execution artifact reports an executor's account of an attempted action. It binds the same Decision and instruction identity and carries an \code{executed\_action\_digest}. The execution event is \code{executed}, \code{failed}, or \code{suppressed}. These outcomes remain distinct.

\subsection{Effect}

An Effect artifact reports an observation of state following a specific Execution. It references both the Decision and the Execution. The observer relationship is declared as \code{same\_executor}, \code{independent}, or \code{unknown}. A same-executor observation is still evidence, but it is not independently corroborated merely by being placed in a separate artifact.

\subsection{Common identity and correlation}

Every family carries \code{airep\_version}, \code{artifact\_type}, \code{chain\_id}, \code{record\_id}, \code{sequence}, \code{subject}, \code{scope}, and \code{integrity}. In the evaluated wire model, \code{airep\_version} is exactly \code{"0.2"}. \code{record\_id} is the stable artifact identity; \code{sequence} is ordering, not identity. Cross-artifact references resolve through record identity, optionally qualified by chain identity; a bare sequence number is not a cross-artifact identifier.

Control and Execution bind the same instruction but distinguish the action authorized from the action executed. Let

\[
D_{auth} = \code{authorized\_action\_digest}, \qquad
D_{exec} = \code{executed\_action\_digest}.
\]

The time-of-check/time-of-use comparison is

\[
D_{auth} = D_{exec},
\]

not equality of either digest with the instruction digest. Equality shows agreement between the recorded authorized and executed action commitments; it does not independently prove execution.

\subsection{Closed core and profiles}

Core objects are closed. Vendor-, model-, framework-, or deployment-specific fields cannot be added arbitrarily to core objects. The \code{profiles} object is the namespaced extension surface. Unknown profiles remain integrity-bound, but without a usable validation basis their profile state is \code{NOT\_EVALUATED}. A profile cannot grant a higher AIREP assurance class by asserting one.

This distinction allows companion evidence to be carried without changing what the four artifact families mean. Section~\ref{sec:integrations} gives an evaluation-evidence example.

\section{Deterministic Wire Processing and Integrity Construction}

Independent recomputation requires implementations to agree before hashing begins. The current wire model therefore specifies both the admitted raw JSON domain and the exact cryptographic preimages.

\subsection{Raw JSON admission}

The processing order is conceptually

\[
\text{raw bytes} \rightarrow \text{admission} \rightarrow \text{JSON model} \rightarrow \text{schema} \rightarrow \text{JCS} \rightarrow \text{integrity}.
\]

A lossy host-language object model cannot precede a check whose evidence that model would destroy. In particular, duplicate member names are rejected recursively before first-wins or last-wins parser behavior can collapse them. AIREP inherits the relevant JCS/I-JSON input constraints selected by RFC 8785~\cite{jcs}, and separately rejects an initial UTF-8 BOM as a deterministic receiver policy. Valid string values are not Unicode-normalized. Non-finite numeric results are rejected before canonicalization.

\subsection{Canonical hash construction}

For artifact $A$ of family $t$, let $B(A)$ be a logical copy of the artifact with only \code{integrity.current} and \code{integrity.signature} removed. All other members, including \code{integrity.previous}, remain. Define

\[
T_H(t) = \operatorname{ASCII}(\text{"AIREP/0.2/hash/"} \Vert t).
\]

The hash preimage is

\[
P_H(A) = T_H(t) \Vert \operatorname{LF} \Vert \operatorname{JCS}(B(A)).
\]

Then

\[
\code{integrity.current} = \text{"sha256:"} \Vert \operatorname{hex}(\operatorname{SHA256}(P_H(A))).
\]

The v0.2 hash contexts are \code{decision}, \code{control}, \code{execution}, and \code{effect}. The registry is closed for wire version 0.2. Tag selection is a function of the artifact's declared version and type; a verifier does not search alternate tags after failure.

\subsection{Signature construction}

The current signature-suite registry contains one suite: pure Ed25519 as defined by RFC 8032~\cite{eddsa}. For artifact type $t$,

\[
T_S(t) = \operatorname{ASCII}(\text{"AIREP/0.2/sig/"} \Vert t),
\]

and the record-signature preimage is

\[
P_S(A) = T_S(t) \Vert \operatorname{LF} \Vert \operatorname{ASCII}(\text{"ed25519"}) \Vert \operatorname{LF} \Vert \operatorname{ASCII}(\code{integrity.current}).
\]

The producer signs $P_S(A)$ directly with Ed25519. Ed25519ph is not the v0.2 suite.

The wire field \code{integrity.signature.alg} is informative only. It is inside the signature object that is removed from the hash preimage, so it is not permitted to select verification behavior. The verifier obtains the suite and public key from its accepted external binding, constructs the preimage with that suite's canonical identifier, and verifies. A mismatch in the wire label can produce a caveat but does not trigger algorithm search.

\subsection{Chains and witness heads}

\code{integrity.previous} binds the current artifact to the preceding artifact's \code{integrity.current}. Genesis uses the defined zero digest. Chain identity, record identity, sequence, version, and artifact type are all inside integrity-protected content.

A partial chain is not proof of a complete chain. AIREP-Witnessed therefore uses a distinct signed head claim containing exactly the chain identity, head sequence, head digest, chain length, and \code{witnessed\_at}. The freshness timestamp used for witness evaluation comes from this signed claim, not from an unsigned auxiliary field.

\section{Assurance Semantics}

AIREP has three assurance classes in the current specification. Table~\ref{tab:classes} states their bounded meanings.

\begin{table}[ht]
\centering
\small
\caption{AIREP assurance classes in the current evaluated specification.}
\label{tab:classes}
\begin{tabularx}{\textwidth}{@{}p{0.23\textwidth}X@{}}
\toprule
\textbf{Class} & \textbf{Establishes, and nothing more} \\
\midrule
AIREP-Core & Accepted structural validity and internal tagged-hash consistency; neither provenance nor freshness \\
AIREP-Authenticated & Core plus authorship under a verifier-accepted producer key binding and current revocation policy \\
AIREP-Witnessed & Authenticated plus independent signed chain-head anchoring, freshness, and non-truncation relative to that accepted anchor \\
\bottomrule
\end{tabularx}
\end{table}

\subsection{Core}

Core is intentionally weaker than tamper-evidence against a substituting adversary. An attacker can fabricate a new, self-consistent Core chain. Core therefore does not establish authorship, freshness, or truth.

\subsection{Authenticated}

Authenticated requires a verifier-accepted binding. A public key self-declared by the producer inside a record cannot grant Authenticated status by itself. The current revocation rule uses the verifier's operator-side binding state. If the required producer binding is revoked, the class ceiling is Core. A producer-declared timestamp cannot restore historical validity to a currently revoked binding; historical validation would require an independently authenticated time basis outside the class definition.

\subsection{Witnessed}

Witnessed requires Authenticated plus an accepted independent witness relation. Independence requires jointly: distinct verifier-accepted identities, different resolved public keys, and explicit verifier policy accepting the witness as independent of the producer. Key inequality alone is insufficient because one actor can control multiple keys.

Witnessed is scoped to the anchor. It does not establish that the evidence graph or real-world history is complete; it establishes freshness and non-truncation relative to the chain head the accepted witness actually vouched for.

\subsection{Failure, withheld, and caveat}

AIREP keeps three outcomes distinct:

\begin{table}[ht]
\centering
\small
\begin{tabularx}{\textwidth}{@{}p{0.19\textwidth}X@{}}
\toprule
\textbf{Outcome} & \textbf{Meaning} \\
\midrule
\code{FAILURE} & A gate was evaluated and definitively did not hold \\
\code{WITHHELD} & A required gate could not be evaluated because necessary operator input was missing or malformed \\
\code{CAVEAT} & A warning attached to an earned class without lowering it \\
\bottomrule
\end{tabularx}
\end{table}

An unevaluated prerequisite is never a pass. If Authenticated cannot be evaluated, the artifact remains at Core and the missing gates are named; similarly, an unevaluable Witnessed tier leaves an otherwise Authenticated artifact at Authenticated.

\section{Lifecycle Reconciliation}

The reconciler evaluates a finite supplied evidence set. It does not claim to reconstruct the world or a complete event history.

Before contributing lifecycle facts, an artifact must pass raw-input admission, its family schema, and the frozen hash check. Invalid artifacts remain visible as admission failures. Per-record class results remain separately visible; an unsigned or unauthenticated assertion does not become authenticated simply because it participates in a successful correlation.

\subsection{Reference and chain resolution}

Record identity must resolve exactly and uniquely. A chain qualifier may constrain a reference but cannot make a duplicated global identity unique. Unresolved identity is missing evidence; wrong-family or mismatching-chain resolution is a contradiction; ambiguous identity is indeterminate. Input order is not allowed to choose a winner.

For supplied chains, sequence ordering and predecessor links are evaluated independently of any claim that the supplied prefix or suffix is complete.

\subsection{Delivery and execution facts}

Controls and Executions are grouped by resolved Decision identity and \code{instruction\_id}. An issuer-side \code{dispatched} report and a receiver-side \code{received} report are different facts. \code{delivery\_failed} is explicit negative evidence; absence of a receiver artifact is not inferred to mean failure.

Only \code{execution\_event=executed} reports execution. \code{failed} and \code{suppressed} remain named negative outcomes. If incompatible outcomes exist without an attempt model or sufficient chronology, the result remains indeterminate.

Authorization-to-execution equality compares applicable Control \code{authorized\_action\_digest} values with Execution \code{executed\_action\_digest} values. Missing Control or Execution evidence prevents the comparison rather than producing an implicit success.

\subsection{Effect coverage and observer independence}

Every Effect resolves to a specific Execution and the same Decision. Effect presence is evaluated per Execution, so evidence for one execution cannot hide absence for another. A declared independent observer becomes effectively independent only if the corresponding authenticated identities, keys, and verifier policy satisfy the independence gate; otherwise the effective relationship remains unknown.

\subsection{Result states}

Every reconciliation check returns exactly one of the states in Table~\ref{tab:reconcile}.

\begin{table}[ht]
\centering
\small
\caption{Reconciliation states.}
\label{tab:reconcile}
\begin{tabularx}{\textwidth}{@{}p{0.20\textwidth}X@{}}
\toprule
\textbf{State} & \textbf{Meaning} \\
\midrule
\code{SATISFIED} & The named predicate holds over the admitted supplied evidence \\
\code{FAILURE} & An evaluated predicate fails or explicit negative evidence reports failure \\
\code{MISSING} & Required evidence is absent from the supplied set \\
\code{NOT\_EVALUATED} & An invalid, unresolved, or ambiguous prerequisite prevents evaluation \\
\code{INDETERMINATE} & Available information does not determine one outcome \\
\bottomrule
\end{tabularx}
\end{table}

All checks are retained. The summary is \code{FAILURE} if any failure exists; otherwise \code{INCOMPLETE} if any missing, unevaluated, or indeterminate check exists; otherwise \code{SATISFIED}. This summary is not a fourth assurance class.

Multi-instruction decisions are evaluated group by group. When no explicit intended-target inventory is supplied, total intended-target coverage remains \code{NOT\_EVALUATED}; a collection of locally complete groups does not prove that all intended targets were represented. This distinction is aligned with the format-independent control-delivery requirements described in the companion Internet-Draft discussed in Section~\ref{sec:related}.

\section{Reference Implementation and Reproducibility}

The released \code{v0.2.0-beta.1} implementation provides a first-party Python producer for all four artifact families, command-line emission tooling, a runnable Decision--Control--Execution--Effect example, Python and Node reference-verification paths, raw-input admission, profile-basis evaluation, structured reconciliation, committed positive and negative fixtures, and a CI-equivalent release runner~\cite{beta}.

The final recorded pre-tag local release command was:

\begin{verbatim}
python3 scripts/check_beta.py --with-v01-typescript \
  --out reports/beta-2026-09-08/release-validation
\end{verbatim}

It was executed under Python 3.12.3 and Node 20.19.6. The release record reports 54 of 54 commands succeeded. Selected measured components are shown in Table~\ref{tab:beta}.

\begin{table}[ht]
\centering
\small
\caption{Selected recorded validation results for \code{v0.2.0-beta.1}.}
\label{tab:beta}
\begin{tabularx}{\textwidth}{@{}X p{0.34\textwidth}@{}}
\toprule
\textbf{Measured surface} & \textbf{Recorded result} \\
\midrule
Complete release command ledger & 54/54 commands succeeded \\
v0.2 producer/reconciliation/admission/profile suite & 39 tests passed \\
Schema engines & 117 fixtures each; comparison/gates passed \\
Historical class-verifier regression corpus & 60 cases; beta adapters matched historical expectations \\
v0.1 regression preservation & 124 pytest tests passed; standalone checks preserved \\
Four-family producer to reference-verifier round trips & Passed \\
Integrity/domain/version/key/tamper negative cases & Passed in the release runner \\
Document preservation audit & 1,919 historical files unchanged in the recorded audit \\
\bottomrule
\end{tabularx}
\end{table}

The release runner also exercised imported W1 contract material, recorded as 423 Python tests with no failures or errors and 2,329 Node assertions passed, with one explicitly unmeasured platform-specific complementary branch on the Python side. These counts are reported because they are part of the recorded release basis, not because they establish independent interoperability.

The 54-command result is explicitly a local CI-equivalent measurement. The publication process separately attached hosted-CI and archive-evidence artifacts to the GitHub prerelease. Same-maintainer Python/Node agreement is not presented as independent third-party implementation evidence.

Post-release integration work described next is pinned to the later manuscript snapshot~\cite{snapshot}. It is not retroactively counted as part of the beta.1 release validation.

\section{Integration Exercises and Extension Surface}\label{sec:integrations}

The repository snapshot used by this manuscript contains two first-party, non-normative integration exercises with external systems. Their purpose is to test whether AIREP's evidence boundaries can be applied without silently changing the semantics of the external system or the AIREP core. They are not claims of external adoption, endorsement, or independent interoperability.

\subsection{Hermes approval/runtime mapping}

A first-party, source-pinned mapping was constructed and tested against the public Nous Research Hermes Agent source snapshot at commit \sha{9796235822b89e08597a402dad045b5b4464e474}~\cite{hermes,hermesintegration}. The mapping treats Hermes as the authorization/enforcement owner and AIREP as an evidence plane: AIREP may report a Decision, Control, Execution, and optional Effect only when the corresponding native fact is available.

The integration corpus contains eleven deterministic, synthetic scenarios and 31 generated AIREP artifacts covering explicit human denial, fail-closed timeout, release without dispatch, dispatch without execution, successful execution, suppression, authorized/executed action mismatch, stale approval response, consumed-but-unobserved execution, a concurrent replay case, and observed effect~\cite{hermesintegration}. The fixtures intentionally preserve non-equivalences that are easy to lose in generic audit logs:

\begin{itemize}
  \item an approval \code{request\_id} is not assumed to be an AIREP Control \code{instruction\_id};
  \item a presentation/request digest is not automatically an AIREP input, instruction, authorized-action, or executed-action digest;
  \item approval or release does not establish instruction dispatch;
  \item dispatch does not establish receiver receipt or execution;
  \item missing execution evidence is not converted into an invented \code{Execution(unknown)} state; and
  \item an effect observation is never inferred to be independent merely because it is represented separately.
\end{itemize}

The generated fixture records reach AIREP-Authenticated only under committed test-only bindings and revocation inputs. That classification authenticates the fixture producer binding/signature under those operator inputs; it does not establish Hermes enforcement correctness, exactly-once consumption, replay prevention, approver authentication, complete history, or Nous Research endorsement. The exercise is therefore first-party integration evidence against an external runtime/source model, not independent Hermes adoption or interoperability.

\subsection{LightEval and the Embedded Evaluation Profile}

The repository also contains an experimental companion profile, \code{airep.embedded-evaluation} version 0.1, carried on AIREP wire 0.2~\cite{snapshot}. The profile is an evaluation-evidence contract, not a fifth artifact family and not a new assurance class. It records declared evaluator/engagement context, target identity, access, evaluation configuration, measurement state, content-addressed evidence, disclosure information, and later verification references while retaining the native evaluation files as source evidence.

A non-normative exporter currently parses Hugging Face LightEval \code{results\_*.json} output~\cite{lighteval}. It hashes the native result file and any supplied details, raw-output, or log files; produces the companion-profile payload, an evidence manifest, and a schema-validation report; and requires a separate context file for facts it must not infer. In particular, evaluator independence, access tier, environment, safeguards, scope, and timezone-aware run timestamps are declared rather than reconstructed from filenames or host state.

The profile keeps execution state separate from observed result. A measurement declared \code{NOT\_RUN} cannot become \code{PASS}; missing evaluation evidence remains \code{NOT\_MEASURED} rather than success. A platform-specific verification token can be hashed and preserved as platform evidence without being reinterpreted as proof that a model is safe, aligned, independently evaluated, or AIREP-Authenticated.

The exporter does not emit a Decision, Control, Execution, or Effect merely to package benchmark output. An AIREP lifecycle artifact is appropriate only where that family's semantics are independently satisfied. This exercise demonstrates the intended role of the namespaced extension surface: companion evidence can be integrity-bound without redefining the four-family lifecycle model.

\section{Independent Implementation Evidence}

The public evidence ledger is version- and role-specific~\cite{ledger}. Results measured against different frozen AIREP versions are reported separately and are not additive.

\subsection{Independent producer evidence on v0.1.2}

Emek Can Do\u{g}ru independently authored a producer against frozen AIREP v0.1.2; the primary implementation and methodology record are published at a pinned repository revision~\cite{emekproducer}. Maintainer-side reproduction recorded in the AIREP evidence ledger found that its output was accepted on first invocation by both pinned v0.1.2 reference verifiers~\cite{ledger}. The authorship-independence claim remains at methodology-and-attestation strength: artifacts can reproduce behavior but cannot prove what source material an implementer did not read.

The exercise also demonstrated that v0.1 admitted more than one plausible interpretation of the record-signature bytes and signature-value encoding. Alternative plausible readings failed the frozen verifier implementations~\cite{emekproducer}. This finding occurred after the v0.2 construction had already pinned the exact signature preimage and wire encoding; it therefore independently confirms that the historical v0.1 ambiguity was real and reachable rather than serving as the design motivation for v0.2. This is historical evidence about the v0.1 line, not v0.2 producer evidence.

\subsection{Independent v0.2 consumer/verifier evidence}

Joel Hillier (Certisyn, Inc.) independently implemented a v0.2 consumer/verifier against the frozen ``AIREP v0.2 Independent-Verifier Corpus v0.1.'' His run record and frozen implementation identity are publicly deposited~\cite{joelverifier}. Maintainer-side reproduction recorded in the AIREP evidence ledger verified the package, reproduced the raw verdict output byte-exact, reproduced six fixed cryptographic vectors byte-exact, and validated all 18 report rows~\cite{ledger}.

The frozen result remains 17 AGREE / 1 DISAGREE. The one disagreement concerns the handoff package's derived \code{cryptographic\_result} expectation for case \code{CLS-XT1}: the frozen expected projection was \code{PASS}, while the independent implementation reported \code{NOT\_EVALUATED} because signature evaluation did not execute after a definitive revoked-binding prerequisite. The evidence ledger records this as an expected-projection defect rather than a failure of the frozen class-verifier contract. Expected outcomes were present in the handoff package, so this is not an expected-blind validation.

This evidence establishes a scoped external consumer/verifier result. It does not establish a third-party v0.2 producer, deployment interoperability, semantic correctness of the protocol as a whole, or standards-body endorsement.

\subsection{Maturity boundary}

The producer result targets v0.1.2; the consumer/verifier result targets a frozen v0.2 corpus. They therefore cannot be combined into one producer-to-consumer interoperability result. At the manuscript snapshot, AIREP does not have a qualifying non-maintainer v0.2 producer and a qualifying non-maintainer v0.2 consumer/verifier measured together against the same frozen candidate. The repository's stable-release criteria retain that external-independence requirement~\cite{stages}.

\section{Relationship to Adjacent Work}\label{sec:related}

This section is deliberately selective. It emphasizes stable standards, peer-reviewed work, and one directly related companion requirements draft rather than attempting an exhaustive catalog of rapidly changing individual Internet-Drafts or agent-observability tools.

\subsection{Model documentation, observability, and authorization}

Model Cards are a model-reporting framework intended to document performance characteristics, evaluation context, and limitations of released models~\cite{modelcards}. They operate at a different granularity from per-decision runtime evidence.

OpenTelemetry Semantic Conventions define common semantic attributes, span names, metric instruments, units, event names, and related telemetry vocabulary~\cite{otel}. Its development-stage GenAI conventions extend this observability model to model, tool, workflow, memory, and agent operations~\cite{otelgenai}. These conventions are close to AIREP in runtime granularity, but they remain telemetry semantics: a span or event does not automatically establish verifier-accepted authorship, governance authority, delivery, execution, effect, or an AIREP assurance class.

OpenID AuthZEN Authorization API 1.0, approved as an OpenID Final Specification in January 2026, standardizes communication of authorization requests and decisions between PEPs and PDPs~\cite{authzen}. AIREP does not replace authorization. An authorization decision can instead be referenced as evidence; AIREP then asks what separately recorded evidence exists for control delivery, execution, and observed effect.

\subsection{Provenance models and secure audit logging}

W3C PROV-DM provides a domain-agnostic model for describing entities, activities, agents, derivations, and responsibility in provenance records~\cite{provdm}. It is useful for provenance interchange and can describe relationships that surround an AIREP record, but it does not prescribe AIREP's closed per-decision artifact families, byte-level cryptographic construction, or assurance ladder.

Secure audit-logging research has long treated the log itself as an adversarial target. Schneier and Kelsey, for example, study cryptographic protection of audit records stored on an untrusted machine under later compromise~\cite{securelogging}. AIREP addresses a different layer: portable event semantics, cryptographic binding, verifier-controlled provenance, and cross-artifact reconciliation. It does not replace host hardening, confidential log storage, or a production secure-logging subsystem.

\subsection{Transparency and supply-chain provenance}

in-toto provides cryptographically verifiable software supply-chain step provenance~\cite{intoto}. RFC 9943 specifies the SCITT architecture for single-issuer signed-statement transparency, including registration with transparency services and receipt-based auditability~\cite{scitt}. Certificate Transparency version 2.0 provides a public-log design for auditable certificate issuance and log behavior~\cite{ctv2}, while Sigstore combines software signing with transparency infrastructure for software-supply-chain use~\cite{sigstore}. AIREP Core does not require a transparency service. AIREP-Witnessed can rely on an accepted witness or transparency anchor, but it does not itself provide a globally consistent public log, gossip, or fork-detection protocol. These mechanisms are complementary rather than interchangeable.

\subsection{Format-independent control-delivery requirements}

A separate individual Internet-Draft by the author, \emph{Evidence Requirements for Agent Control Delivery and Outcome Reconciliation}, revision -01, defines format-independent requirements that separate issuer-side emission, required-target resolution, receiver-side observation, enforcement outcome, and observation of the resulting control effect~\cite{control-draft}. It is a Work in Progress and does not represent IETF endorsement. The draft deliberately does not define a receipt format. AIREP is one concrete protocol with overlapping concerns; the draft is not an AIREP conformance authority and includes bounded-population accounting requirements that the current beta reconciler does not fully instantiate.

\subsection{Governance frameworks}

NIST AI RMF 1.0 is a voluntary, non-sector-specific risk-management framework~\cite{nist}. Article 12 of Regulation (EU) 2024/1689 requires high-risk AI systems to technically allow automatic event logging over the system lifetime and ties logging capabilities to traceability and monitoring purposes~\cite{euai}. AIREP can produce technical runtime evidence relevant to audit and traceability workflows, but it does not by itself establish compliance with either framework or with any legal obligation.

\section{Limitations}

\textbf{Malicious producers.} A valid signature does not make a producer honest. A key holder can emit a correctly signed false report.

\textbf{External trust roots.} AIREP-Authenticated depends on verifier-accepted key bindings and revocation inputs. General organizational identity, key issuance, delegation, and trust-store governance remain external.

\textbf{Anchor-relative witnessing.} Witnessed establishes freshness and non-truncation relative to what an accepted witness anchored. It does not prove complete visibility into every relevant event or chain.

\textbf{Compromised or equivocating witnesses.} Verifier-accepted independence is not a guarantee of witness honesty. A compromised, colluding, or equivocating witness can sign conflicting chain heads. AIREP-Witnessed is therefore relative to the accepted anchor presented to the verifier and does not, by itself, provide global fork consistency. Detecting equivocation across observers requires an external mechanism such as a transparency log with independent auditing/consistency checks, cross-observer gossip, or comparison across multiple independently operated witnesses~\cite{ctv2,sigstore}.

\textbf{Finite evidence sets.} The reconciler evaluates what it is given. No receipt does not prove non-delivery; no Execution does not prove non-execution; no Effect does not prove no effect.

\textbf{Target completeness.} Without an explicit intended-target inventory, total target coverage remains unevaluated.

\textbf{Privacy and confidentiality.} Stable identities and cross-record references improve audit correlation but can create linkability across systems and time. A cryptographic digest is an integrity commitment, not an encryption mechanism; a digest of low-entropy or guessable content can support offline guessing or dictionary tests. Deployments should minimize personal or sensitive data in artifacts, separate large or sensitive source content from content-addressed references where appropriate, enforce access control and encryption for retained evidence, and define retention/erasure policies consistent with their legal and operational obligations. Selective-disclosure, salted, or keyed commitment schemes may be appropriate outside the current Core, but they should not be silently substituted for normative AIREP SHA-256 fields because that would change what the wire digest means. AIREP Core does not itself provide confidentiality, unlinkability, deletion, or privacy compliance.

\textbf{Exact historical policy replay.} The current Core records policy basis and can bind supplied policy evidence, but it does not require a version-pinned policy snapshot sufficient to reconstruct every historical policy dependency. Policy-correctness replay therefore requires additional deployment-specific evidence or profile semantics.

\textbf{Production key management.} The beta exposes reference development key/signing surfaces, not a production HSM/KMS deployment architecture or concurrent ledger manager.

\textbf{Hardware attestation.} TEE and hardware attestation are outside the current Core. They may be referenced as external evidence or profiles without becoming Core guarantees.

\textbf{Security review.} The project publishes adversarial tests and a vulnerability-reporting policy, but no independent security certification or comprehensive third-party security audit is claimed.

\textbf{Performance, scalability, and usability.} The measurements reported here are correctness, conformance, and reproducibility oriented. They do not measure sustained throughput, latency, storage growth, multi-writer contention, large-scale reconciliation cost, operator usability, or production deployment reliability.

\textbf{Integration generality.} The Hermes exercise is a first-party, source-pinned mapping with synthetic fixtures; the LightEval exporter is a first-party non-normative integration that parses one native result shape. Neither establishes broad framework interoperability or external adoption.

\textbf{Migration.} v0.1 artifacts remain governed by v0.1 rules. They are not rehashed, resigned, or relabelled as v0.2. A migration projector is not part of the beta release.

\textbf{Interoperability maturity.} The principal open release-level requirement is same-candidate external interoperability. Local green tests, first-party integrations, and same-maintainer cross-language agreement cannot satisfy the non-maintainer stable-release gate.

\section{Discussion}

AIREP's most consequential design choice is semantic rather than cryptographic: it refuses to collapse evidentiary stages. Operational statements such as ``the system blocked it,'' ``the stop was sent,'' ``the agent was stopped,'' and ``the effect was observed'' can sound equivalent in ordinary language while denoting different evidence boundaries.

The four-family model turns that ambiguity into explicit structure. A Decision is a first-class governance-decision artifact. A receiver-side Control can report receipt. An Execution is an executor report. An Effect is an observation. The assurance ladder then answers a different question: what provenance, integrity, and freshness properties does a particular record earn? Reconciliation answers a third: how do the supplied records relate to one another?

Keeping these axes separate---event type, record assurance, and cross-record reconciliation---prevents an increase in one from being silently interpreted as an increase in another. Stable record identity and explicit cross-artifact references make the decision usable as a join point without requiring the protocol to impose one server architecture or one runtime-specific execution identifier.

This distinction becomes more important in agentic systems because authority and action frequently cross boundaries. The component proposing an action may differ from the component authorizing it; the authorizer may differ from the dispatcher; the enforcement point may be remote; the executor may be operated by another party; and the observer may or may not be independent of the executor. A generic audit event can obscure these boundaries. AIREP instead requires the evidence to name them when known and to preserve missing, withheld, unevaluated, or indeterminate states when they are not established.

The integration exercises reinforce the same point from two different directions. The Hermes mapping shows that approval, dispatch, receipt, execution, and effect cannot safely be collapsed even when one runtime provides several related identifiers and states. The LightEval integration shows that companion evaluation evidence can be preserved through the extension surface without manufacturing a Decision--Control--Execution--Effect lifecycle that did not occur. Both exercises are useful precisely because they preserve non-equivalence rather than forcing external systems into an AIREP-shaped success narrative.

\section{Reproducibility and Artifact Availability}

The released implementation basis described in Section 7 is \code{v0.2.0-beta.1}, published 9 September 2026. The annotated tag resolves to release commit

\begin{center}
\sha{8a6c01ecce457aa94330c0ed7219e4c56ebfe771}.
\end{center}

The release page and source archives are available from the AIREP repository~\cite{beta}. The project concept DOI is \href{https://doi.org/10.5281/zenodo.20475136}{10.5281/zenodo.20475136}.

The post-release evidence and integration statements in Sections 8 and 9 are pinned for this manuscript to repository commit

\begin{center}
\sha{32f2ded6431025bf2737a7abe7152f24ccf72679},
\end{center}

which contains the LightEval/Embedded Evaluation integration, the source-pinned Hermes mapping and deterministic fixtures, and the external-evidence ledger used here~\cite{snapshot,ledger}. Pinning the released implementation and the later manuscript evidence snapshot separately avoids retroactively changing what the beta tag itself contained.

The frozen v0.1 line, v0.2 schemas, integrity specification, reconciler contract, class-verifier corpora, integration fixtures, and evidence ledger are all publicly inspectable in the repository at the cited commits. The v1 manuscript remains available as arXiv:2608.21363v1~\cite{airepv1}.

\section{Conclusion}

AIREP represents AI runtime-governance evidence as four explicitly typed artifact families: Decision, Control, Execution, and Effect. The current evaluated wire instantiation combines closed-schema records, explicit references and digests, scope declarations, deterministic raw-input admission, RFC 8785 canonicalization, domain-separated SHA-256 hashing, pure Ed25519 signatures, chain linkage, verifier-controlled trust bindings, independent witness semantics, and structured lifecycle reconciliation.

The design deliberately separates three questions:

\begin{enumerate}
  \item What lifecycle event does this record report?
  \item What provenance, integrity, and freshness assurance does the record earn?
  \item How does the record reconcile with the other evidence supplied?
\end{enumerate}

None of these answers establishes the truth of the underlying AI output or a complete real-world history. The protocol's extension surface likewise allows companion evidence to be carried without silently changing lifecycle semantics or assurance classes.

The released beta demonstrates a runnable first-party four-family reference implementation and a reproducible recorded release-validation run. Post-release Hermes and LightEval exercises provide first-party integration evidence against external source/evaluation models while preserving explicit non-equivalences. Independent implementation evidence exists in both producer and consumer/verifier roles across the AIREP project, but the measurements target different frozen version bases. They therefore do not yet establish the same-version independent interoperability required by the protocol's own stable-release criteria.

That boundary is part of the contribution rather than a disclaimer appended to it. An evidence protocol should not claim more than its evidence establishes. AIREP's purpose is not to make AI claims true; it is to make mechanically inspectable what was claimed, by whom, about which lifecycle stage, under which cryptographic and trust conditions, and where the available evidence ends.

\begingroup\footnotesize
\setlength{\parskip}{0.1em}

\endgroup


\begin{thebibliography}{99}

\bibitem{modelcards}
M. Mitchell, S. Wu, A. Zaldivar, P. Barnes, L. Vasserman, B. Hutchinson, E. Spitzer, I. D. Raji, and T. Gebru, ``Model Cards for Model Reporting,'' \emph{Proceedings of the Conference on Fairness, Accountability, and Transparency}, pp. 220--229, 2019. doi: \href{https://doi.org/10.1145/3287560.3287596}{10.1145/3287560.3287596}.

\bibitem{otel}
OpenTelemetry Authors, \emph{OpenTelemetry Semantic Conventions 1.44.0}, 2026. Version-pinned source: \url{https://github.com/open-telemetry/semantic-conventions/tree/v1.44.0}.

\bibitem{otelgenai}
OpenTelemetry Authors, \emph{Semantic Conventions for Generative AI Systems}, development-stage source snapshot commit \sha{c88d504ab3d9879f8e50d3cc87e69775e11db234}, 16 September 2026. \url{https://github.com/open-telemetry/semantic-conventions-genai/tree/c88d504ab3d9879f8e50d3cc87e69775e11db234/docs/gen-ai}.

\bibitem{authzen}
O. Gazitt, D. Brossard, and A. Tulshibagwale, eds., \emph{Authorization API 1.0}, OpenID Foundation AuthZEN Working Group, Final Specification, 11 January 2026. \url{https://openid.net/specs/authorization-api-1_0.html}.

\bibitem{intoto}
S. Torres-Arias, H. Afzali, T. K. Kuppusamy, R. Curtmola, and J. Cappos, ``in-toto: Providing farm-to-table guarantees for bits and bytes,'' \emph{28th USENIX Security Symposium (USENIX Security 19)}, pp. 1393--1410, 2019. \url{https://www.usenix.org/conference/usenixsecurity19/presentation/torres-arias}.

\bibitem{scitt}
H. Birkholz, A. Delignat-Lavaud, C. Fournet, Y. Deshpande, and S. Lasker, \emph{An Architecture for Trustworthy and Transparent Digital Supply Chains}, RFC 9943, IETF, June 2026. doi: \href{https://doi.org/10.17487/RFC9943}{10.17487/RFC9943}.

\bibitem{provdm}
L. Moreau and P. Missier, eds., \emph{PROV-DM: The PROV Data Model}, W3C Recommendation, 30 April 2013. \url{https://www.w3.org/TR/2013/REC-prov-dm-20130430/}.

\bibitem{securelogging}
B. Schneier and J. Kelsey, ``Secure Audit Logs to Support Computer Forensics,'' \emph{ACM Transactions on Information and System Security}, vol. 2, no. 2, pp. 159--176, 1999. doi: \href{https://doi.org/10.1145/317087.317089}{10.1145/317087.317089}.

\bibitem{ctv2}
B. Laurie, E. Messeri, and R. Stradling, \emph{Certificate Transparency Version 2.0}, RFC 9162, IETF, December 2021. doi: \href{https://doi.org/10.17487/RFC9162}{10.17487/RFC9162}.

\bibitem{sigstore}
Z. Newman, J. S. Meyers, and S. Torres-Arias, ``Sigstore: Software Signing for Everybody,'' \emph{Proceedings of the 2022 ACM SIGSAC Conference on Computer and Communications Security}, pp. 2353--2367, 2022. doi: \href{https://doi.org/10.1145/3548606.3560596}{10.1145/3548606.3560596}.

\bibitem{nist}
E. Tabassi, \emph{Artificial Intelligence Risk Management Framework (AI RMF 1.0)}, NIST AI 100-1, National Institute of Standards and Technology, January 2023. doi: \href{https://doi.org/10.6028/NIST.AI.100-1}{10.6028/NIST.AI.100-1}.

\bibitem{euai}
European Parliament and Council of the European Union, \emph{Regulation (EU) 2024/1689 laying down harmonised rules on artificial intelligence (Artificial Intelligence Act)}, consolidated text of 27 July 2026, Article 12 (Record-keeping). \url{https://eur-lex.europa.eu/legal-content/EN/TXT/?uri=CELEX:02024R1689-20260727}.

\bibitem{jcs}
A. Rundgren, B. Jordan, and S. Erdtman, \emph{JSON Canonicalization Scheme (JCS)}, RFC 8785, June 2020. doi: \href{https://doi.org/10.17487/RFC8785}{10.17487/RFC8785}.

\bibitem{eddsa}
S. Josefsson and I. Liusvaara, \emph{Edwards-Curve Digital Signature Algorithm (EdDSA)}, RFC 8032, January 2017. doi: \href{https://doi.org/10.17487/RFC8032}{10.17487/RFC8032}.

\bibitem{lighteval}
N. Habib, C. Fourrier, H. Kydl\'i\v{c}ek, T. Wolf, and L. Tunstall, \emph{LightEval: A lightweight framework for LLM evaluation}, Hugging Face, software repository, version 0.11.0. Version-pinned source: \url{https://github.com/huggingface/lighteval/tree/v0.11.0}.

\bibitem{hermes}
Nous Research, \emph{Hermes Agent}, public source repository, source snapshot commit \sha{9796235822b89e08597a402dad045b5b4464e474}, 16 September 2026. \url{https://github.com/NousResearch/hermes-agent/tree/9796235822b89e08597a402dad045b5b4464e474}.

\bibitem{hermesintegration}
A. T. Abak, \emph{Hermes approval/runtime to AIREP v0.2 mapping and deterministic fixture corpus}, AI Runtime Evidence Protocol repository, mapping version 0.1, manuscript snapshot commit \sha{32f2ded6431025bf2737a7abe7152f24ccf72679}, 16 September 2026. \url{https://github.com/halvrenofviryel/ai-runtime-evidence-protocol/tree/32f2ded6431025bf2737a7abe7152f24ccf72679/integrations/hermes}.

\bibitem{control-draft}
A. T. Abak, \emph{Evidence Requirements for Agent Control Delivery and Outcome Reconciliation}, Internet-Draft \code{draft-abak-agent-control-delivery-evidence-01}, Work in Progress, 4 September 2026. Revision-pinned text: \url{https://datatracker.ietf.org/doc/html/draft-abak-agent-control-delivery-evidence-01}.

\bibitem{airepv1}
A. T. Abak, ``AIREP: A Protocol for Per-Decision Evidence in AI Runtime Governance,'' arXiv:2608.21363v1, 2026. Version-pinned manuscript: \url{https://arxiv.org/abs/2608.21363v1}. doi: \href{https://doi.org/10.48550/arXiv.2608.21363}{10.48550/arXiv.2608.21363}.

\bibitem{emekproducer}
E. C. Do\u{g}ru, \emph{Independent AIREP v0.1 producer}, public source repository, measured against frozen v0.1.2; pinned commit \sha{31e12be987105d2ba93ab2abe6135d9e6d5374d2}, 2026. \url{https://github.com/dogrucanemek-alt/airep-independent-producer/tree/31e12be987105d2ba93ab2abe6135d9e6d5374d2}.

\bibitem{joelverifier}
J. Hillier, Certisyn, Inc., \emph{AIREP v0.2 Independent-Verifier Corpus v0.1 --- run record}, external artifact, implementation digest \code{sha256:2aef1212adeaab5a1dc7f07c3f240183db97478b247c008c5fcc0e177fbfeca8}, 2026. \url{https://github.com/halvrenofviryel/ai-runtime-evidence-protocol/blob/32f2ded6431025bf2737a7abe7152f24ccf72679/external-evidence/certisyn-2026-09/AIREP-RUN-RECORD.txt}.

\bibitem{snapshot}
A. T. Abak, \emph{AI Runtime Evidence Protocol (AIREP) repository}, manuscript evidence snapshot commit \sha{32f2ded6431025bf2737a7abe7152f24ccf72679}, 16 September 2026. \url{https://github.com/halvrenofviryel/ai-runtime-evidence-protocol/tree/32f2ded6431025bf2737a7abe7152f24ccf72679}.

\bibitem{beta}
A. T. Abak, \emph{AI Runtime Evidence Protocol (AIREP), v0.2.0-beta.1}, GitHub prerelease, 9 September 2026; release commit \sha{8a6c01ecce457aa94330c0ed7219e4c56ebfe771}. Concept DOI: \href{https://doi.org/10.5281/zenodo.20475136}{10.5281/zenodo.20475136}. \url{https://github.com/halvrenofviryel/ai-runtime-evidence-protocol/releases/tag/v0.2.0-beta.1}.

\bibitem{ledger}
A. T. Abak, \emph{AIREP External Evidence Ledger}, AI Runtime Evidence Protocol repository, manuscript snapshot commit \sha{32f2ded6431025bf2737a7abe7152f24ccf72679}, 16 September 2026. \url{https://github.com/halvrenofviryel/ai-runtime-evidence-protocol/blob/32f2ded6431025bf2737a7abe7152f24ccf72679/EXTERNAL_EVIDENCE.md}.

\bibitem{stages}
A. T. Abak, \emph{AIREP v0.2 Release Stages}, AI Runtime Evidence Protocol repository, \code{v0.2.0-beta.1} release basis, 2026. \url{https://github.com/halvrenofviryel/ai-runtime-evidence-protocol/blob/v0.2.0-beta.1/spec/airep/v0.2/RELEASE_STAGES.md}.

\end{thebibliography}
\end{document}